# Considerations on Construction Ontologies

**Alexandru Cicortaş[1], Victoria Stana Iordan[1], Alexandra Emilia Fortiş[2]**
[1]**West University, Timişoara, Romania**
[2]**"Tibiscus" University, Timişoara, Romania**

**ABSTRACT**. The paper proposes an analysis on some existent ontologies, in order to point out ways to resolve semantic heterogeneity in information systems. Authors are highlighting the tasks in a Knowledge Acquisiton System and identifying aspects related to the addition of new information to an intelligent system. A solution is proposed, as a combination of ontology reasoning services and natural languages generation. A multi-agent system will be conceived with an extractor agent, a reasoner agent and a competence management agent.

**Introduction**

The need for increasing the cognitive support knowledge engineering is a real requirement and a visual representation is a real need. Between knowledge representations are: Sowa based on the KIF [Gin91], CODE4 [SL95], focused in more detail on the user experience, and also combined it with a logically rigorous representational semantics.

In many domains like business-to-business e-commerce are required dynamic and open-interoperable information systems that are service-oriented. Services are often made of sub-services and tasks that normally belong to autonomous participants. Inevitably the underlying information systems are distributed and autonomous.

The semantics of diverse information sources is captured by their ontologies, i.e., the terms and relationships between them [CB00]. In tightly coupled applications, the intended meaning of a term is often implicit, thus relying on developer's mutual agreement. In a distributed environment mutual agreement is hard to come by if not impossible. Thus it is crucial for the domain model and the vocabulary to be represented in such a way that enables programs to reuse them as they were originally intended with minimum human intervention during their execution.

Semantic heterogeneities represent another major problem that must be carefully analyzed. Heterogeneity in many domains is inevitable because the concerned systems are often developed by autonomous participants. Semantically equivalent concepts:

- Different terms are used to refer the same concept by two models. These terms are often called synonyms. However, synonyms in their common usage do not necessarily denote semantically equivalent concepts.
- Different properties are modelled by two systems (heterogeneity). As an example, for the same product, one catalogue has included its colour but the other has not.
- Property-type mismatches. For example, the concept length may be given in different units of measure.

Semantically unrelated concepts here the conflicting term are a concept. The same term may be chosen by two systems to denote completely different concepts.

Semantically related concepts are:
- Generalization and specification. As an example is that student in one system refers to all students, but the other only to PhD students.
- Definable terms or abstraction - A term may be missing from one ontology, but which can be defined in other terms in the ontology,
- Overlapping concepts.
- Different conceptualisation. Example: one ontology classifies person as male, female, and the other person as employed unemployed.

Ontology can be seen as a way to resolve semantic heterogeneity by specifying explicitly the semantics of the terms used in information systems. Between multiple ontology definitions the two following seem to be useful for our intentions. Ontology is an explicit specification of a conceptualisation [Gru92]. Ontology is a logical theory accounting for the intended meaning of a formal vocabulary, i.e., its ontological commitment to a particular conceptualisation of the world [Gua98].

**1 Previous Works**

In [All03] was done an analysis of the visualization tools for knowledge engineering, from that the lack of an established theory about user tasks and the cognitive support they require was revealed. Also were identified many difficulties encountered when performing user testing in the knowledge engineering domain, including gaining access to expert users, generalizing results over different domains and quantifying the knowledge acquired and used by such tools.

These issues recommend focusing on more qualitative approaches which included a user survey, two contextual inquiries, and investigation of related work; using these different techniques provided a series of useful perspectives on the problem. There exists a wide variety of users and domains to which ontology engineering is being applied, and further, that visualization is a desired feature.

Based on an analysis of the Knowledge Acquisition system [TKG01] the experimenters observed users performing the following high-level tasks:

- understanding the given knowledge acquisition task;
- deciding how to proceed with the knowledge acquisition task;
- browsing the knowledge base to understand it;
- browsing the knowledge base to find something;
- editing (create or modify) a knowledge base element;
- checking that a modification had the expected effects on the knowledge base;
- looking around for possible errors;
- understanding and deciding how to fix an error;
- recovering from an error by undoing previous steps (to delete or restore a knowledge base element);
- reasoning about the system.

In [BKR01] were identified some typical concerns that users may have when adding new knowledge to an intelligent system. Some of these concerns were that the users do not know where to start and where to go next, the users do not know if they are adding the right things and the users often get lost as it takes several steps to add new knowledge. Here is clearly shoed that the standard knowledge engineering methodology, consisting of the steps of modelling, acquiring and verifying knowledge, fails to accommodate the specific needs of users, even modellers in the domain. It is no use to have a crisp and detailed methodology if users cannot easily make use of it in any one of its stages.

In [Ng00], based on an evaluation of user requirements in ontology modelling tasks, was designed a tool, Info Lens, to browse description logic ontologies (using a combination of lenses) which revealed different information about the domain as they were interactively moved about the model representation. One issue was scalability for practical sized systems. For cognitive support specific tasks, were identified the need for a tool to support information integration (between different representations), to support the often cyclic task-switching between navigation (around the model) and visualization (of a specific aspect of the model). Initial user surveys were quite positive but some aspects of the implementation hindered the evaluation. In [C+01] is described a graphical tool for knowledge acquisition. Although they only tested it on four users, and those users were not modellers or knowledge engineers, we still present the results for the insight it offers into the benefit of increased cognitive support. The users were able to enter a few hundred concepts into a large medical knowledge base within a week, and also verify the model using competency questions. As major problems were extracted the basic machinery works, providing a basic vehicle for axiom-building without the users having to encode axioms directly or even encounter terms like concept, relation, instance, quantification.

Also were identified as areas needing improvement include multifaceted representations, active critiques from the system and more expressivity in the interface such as temporal relations and conditions.

Protege is an ontology engineering and knowledge acquisition tool created at Stanford University [G+03]. It uses a frame-based knowledge representation formalism to allow users to model domains using classes, instance, slots (relations) and facets (constraints on the slots). Written in Java, its architecture allows for extensions to be added via a plug-in metaphor. Recently, work has been ongoing to make the tool compatible with the OWL ontology language for the Semantic Web (a key component), as well as support web-specific concepts such as namespaces and Universal Resource Indicators (URIs). More details are available in [KMN03].

## 2 Managing ontologies

Ontology is seen as domain oriented concepts. It includes abstract concepts and specifies domain-level constraints that can be used for knowledge-level reasoning; Ontology is suited to represent high-level information requirements. Schemas and classes are data-level concepts that are implementation dependent. They are designed to optimise procedural operations. Constraints at this level are operational constraints. Many domain constraints are not explicitly represented at this level. The relationship between ontology and the underlying data sources are represented in the following figure:

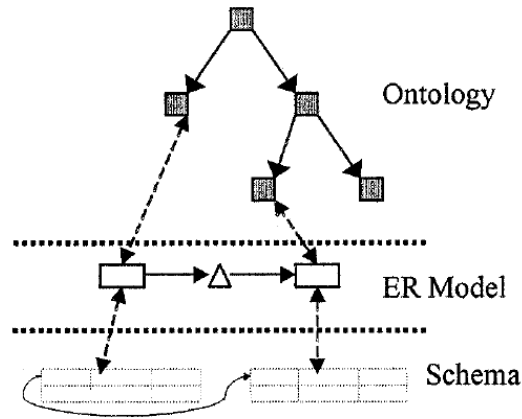

Figure 1: The relationship between ontology

In addition, DOME ontologies form a hierarchy as shown in Figure 2.

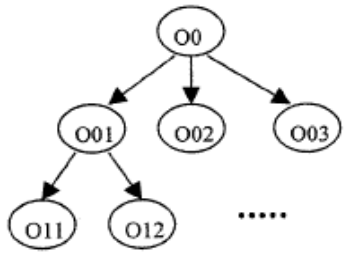

Figure 2: Ontology hierarchy

## 3 Learning ontologies

Besides learning ontologies from existing data sets, we can also reuse existing ontologies available from the Web [SP07]. The first step is to get the candidates by using ontology search tools like OntoSearch [T+04], [JT06]. The major problem consists from finding the way that allows to the ordinary users, who may not know OWL at all, decide which ontology suits their application best. A good solution [P+02] is to combine ontology reasoning services and natural language generation to provide human read able presentation of parts of ontologies. Ontology takes the form of a set of logical axioms, and so the challenge is to present the material of these axioms in comprehensible way using a language such as English. However, it is important to take on board the fact that the axioms may not come in a form ready for direct realisation in English. The axioms represent one possible way that the material could have been expressed, but there are many other possible ways that this could have been done equally well.

## 4 Contributions

In the domain we had many papers [CIN08], [IC08] that treats the competence representation and description using ontologies. As we stated above, based on the remarks from [OM06] and in accordance with our intentions, the following is proposed.

The problem statement can be summarized as follows: own the ontology in order to construct a tool that generates another ontology based on appropriate inference and reasoning.

As a simple example in education: having many course descriptions define the skills and the capabilities and based on these, derive the competences that can be obtained attending these courses.

In one of the previous sections were presented the operations on the ontologies. As it can be seen we propose another operation deriving ontology from other one. For that we have at least two possible solutions:
- conceive an expert system with appropriate goals;
- conceive some intelligent agents that are able to do it in an appropriate context.

The multi-agent system has as main goal to derive ontology from another one in the following way. It will extract from course descriptions the possible skills and capabilities. From the skills and capabilities, the competences that are acquired which are expressed in terms of a new ontology.

The system has three agents: Extractor, Reasoner and Competence Management Agents.

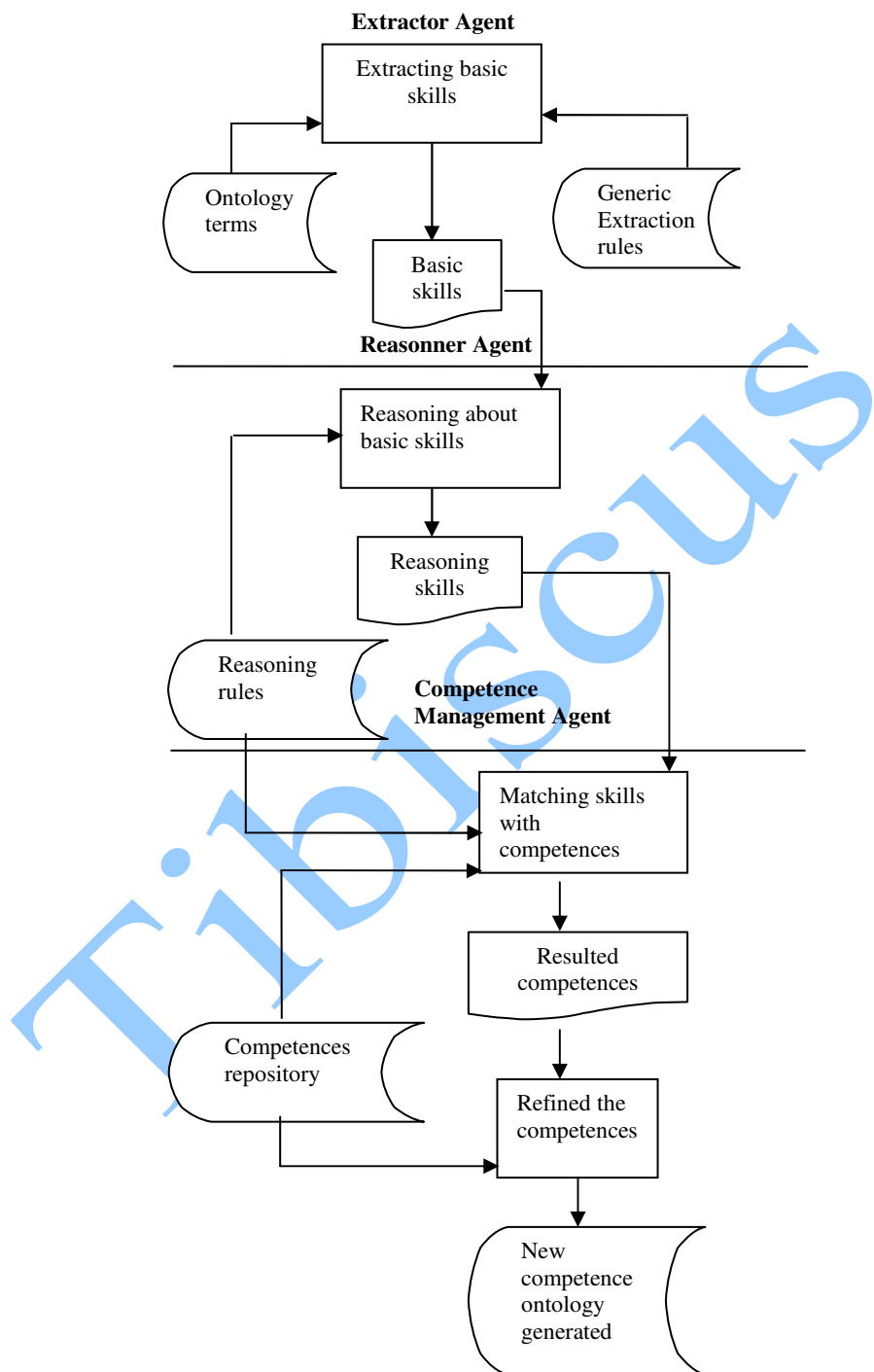

Fig. 3: Multi-Agent system for extracting the competences

As is shown in the figure 3, the ontology that describes the courses and the generic rules are used by the Extractor Agent. It extracts the skills and capabilities that are obtained after attending these courses. After it, the Reasoner Agent defines the possible competences from the skills. These competences are refined based on the comparisons with the similar competences that exist in the Competence repository and the resulted ontology (of the new competences) is obtained. Our model has some similitude and some functionality like the model presented in [L+05]. As basis for information and knowledge representation, the XML will be used. The main motivation is due to the fat that on the Internet the information must be extracted processed and presented in some specific form.

Based on it the agents will be able use the information for communicating each other and with the users.

**Conclusions**

Ontologies are frequently used in design of complex systems especially in the case of agent usage. Due to the similarities between the ontologies and competences, the ontology construction can be also used for competence construction. The multi-agent concepts are used in ontology construction and based on it the different tools was developed.

Future our works will concentrate on the refinement of the agent capabilities for construction the ontologies and competences.